\documentclass[runningheads]{llncs}
\usepackage{graphicx}
\usepackage{amsfonts}
\usepackage{amsmath}
\usepackage{multirow}
\usepackage{url}
\usepackage{xcolor}
\begin{document}
\title{Leveraging Persistent Homology for Differential Diagnosis of Mild Cognitive Impairment}
\author{Ninad Aithal
% \orcidID{0000-1111-2222-3333} 
\and
Debanjali Bhattacharya 
% \orcidID{1111-2222-3333-4444} 
\and
Neelam Sinha 
% \orcidID{2222--3333-4444-5555}
% \email{neelam@cbr-iisc.ac.in}
\and
Thomas Gregor Issac
}
\authorrunning{N. Aithal et al.}
\institute{Center for Brain Research, Indian Institute of Science (IISc), India}

\maketitle  
%%%%%%%%%%%%%%%%%%%%%%%%%%%%%%%%%%%%%%%%%%%%%%%%%%%%%%%%%%%%%%%%%%%%%%%%
\begin{abstract}
Mild cognitive impairment (MCI) is characterized by subtle changes in cognitive functions, often associated with disruptions in brain connectivity. The present study introduces a novel fine-grained analysis to examine topological alterations in neurodegeneration pertaining to six different brain networks of MCI subjects (Early/Late MCI). To achieve this, fMRI time series from two distinct populations are investigated: (i) the publicly accessible ADNI dataset and (ii) our in-house dataset. The study utilizes sliding window embedding to convert each fMRI time series into a sequence of 3-dimensional vectors, facilitating the assessment of changes in regional brain topology. Distinct persistence diagrams are computed for Betti descriptors of dimension-0, 1, and 2.  Wasserstein distance metric is used to quantify differences in topological characteristics. We have examined both (i) ROI-specific inter-subject interactions and (ii) subject-specific inter-ROI  interactions. Further, a new deep learning model is proposed for classification, achieving a maximum classification accuracy of 95\% for the ADNI dataset and 85\% for the in-house dataset. This methodology is further adapted for the differential diagnosis of MCI sub-types, resulting in a peak accuracy of 76.5\%, 91.1\% and 80\% in classifying HC Vs. EMCI, HC Vs. LMCI and EMCI Vs. LMCI, respectively. We showed that the proposed approach surpasses current state-of-the-art techniques designed for classifying MCI and its sub-types using fMRI.

\keywords{fMRI time series  \and Sliding window embedding \and Persistent homology \and Wasserstein distance \and Deep learning.}

\end{abstract}
%%%%%%%%%%%%%%%%%%%%%%%%%%%%%%%%%%%%%%%%%%%%%%%%%%%%%%%%%%%%%%%%

\section{Introduction}

Mild Cognitive Impairment (MCI) stands as a crucial stage bridging normal cognitive aging and dementia, often serving as a precursor to conditions like Alzheimer's disease (AD) and other neurodegenerative disorders \cite{janoutova2015mild,ammu1,ninad}. Research indicates that individuals with MCI progress to AD at a rate of approximately 10–15\% per year \cite{new1}, making MCI as the most challenging group for early detection and diagnosis of AD. Based on the extent of episodic memory impairment, MCI can be primarily categorized into Early Mild Cognitive Impairment (EMCI) and Late Mild Cognitive Impairment (LMCI). Notably, the risk of LMCI transitioning to AD surpasses that of EMCI. However, detecting EMCI remains clinically challenging due to subtle alteration from healthy controls (HC). Additionally, the classification of EMCI and LMCI based solely on memory scores may lead to low specificity and misclassifications. The search for sensitive biomarkers that change alongside disease progression offers hope in refining disease staging, potentially decreasing the prevalence of AD through early intervention. To be more specific, EMCI is particularly an important sub-type of MCI for implementing interventions, aimed at potentially modifying the progression of the condition. Hence, there is a growing emphasis on delineating the neurobiological alterations associated with EMCI and LMCI, vital for early diagnosis, prognosis, and intervention strategies \cite{morley2011anticholinergic}. The identification of potentially high-sensitivity diagnostic markers evolving alongside disease progression can significantly aid physicians in making accurate diagnoses.
Recent improvements in brain imaging, especially with fMRI, offer valuable insights into how MCI affects brain function. By studying changes in various brain networks, we gain a better understanding of how MCI disrupts communication within the brain. These disruptions are key to understanding the disease early on, which could help delay or reverse cognitive decline associated with MCI and its sub-types.
In the current study, rather than relying on state-of-the-art machine learning and deep learning methods which are commonly used for classification of MCI and its sub-types using various MRI modalities\cite{ammu1,ninad}, we have reported completely different approach. We have utilized \textit{persistent homology}- an advanced tool in computational topology, in order to explore potential differences in topology between MCI sub-types and compared them to HC. This novel approach diverges from conventional feature engineering techniques and offers a unique perspective on understanding the subtle yet significant variations associated between MCI sub-types and HC. \par
Persistent homology is a powerful topological data analysis (TDA) approach falling under the branch of algebraic topology. It provides a robust framework for analyzing the topological features of data, particularly in the context of shape and structure. At its core, persistent homology aims to capture the evolution of topological features across different spatial scales by constructing a sequence of topological spaces based on the input data. It reveals how specific topological characteristics remain consistent or evolve as we observe these spaces at different levels of detail. Persistent homology finds its application in various domains. In the context of medical image analysis, it is used for analysis of endoscopy \cite{2016-Dunaeva-Endoscopy-Analysis}, breast cancer \cite{unknown}, analysis of brain networks for differentiating various types of brain disorders \cite{2012-Lee-Persistent-Homology-Brain-Networks,2021-Stolz-schizophrenia-experiments}, detecting transition between states in EEG \cite{2016-Merelli-EEG}, identifying epileptic seizures \cite{2021-Caputi-Epilepsy} and distinguishing between male and female brain networks \cite{2022-Das-TDA-Brain-Networks}. Notably, persistent homology of time series emerges as a rapidly growing field with several compelling applications like computing stability of dynamic systems \cite{ph2}, to quantify periodicity \cite{sw1} and differences in visual brain networks \cite{boldfmrits}. \par
In this study, we analyze fMRI time series data derived from 160 Dosenbach Regions of Interest (ROIs), which are selected from six classical brain networks. Each ROI corresponds to a distinct time series, and our focus is on leveraging persistent homology to analyze these fMRI time series for the classification of MCI and its sub-types.
%Our findings reveal notable changes in brain network topology between HC and MCI sub-types. These changes are characterized by shifts in the persistence profiles of topological features like connected components, loops, and voids. Such alterations suggest disruptions in functional connectivity patterns and network organization linked to MCI pathology, reflecting the underlying neurodegenerative processes and cognitive decline. Additionally, we conduct statistical analysis and utilize deep learning based classification to quantitatively assess the discriminatory power of persistent homology-derived features in distinguishing between HC and MCI sub-types. Our results underscore the effectiveness of persistent homology in capturing subtle significant differences in brain network topology, thereby providing valuable insights into the neuropathology of MCI and its sub-types.
The present study offers two significant contributions: \\
(i) The introduction of a novel paradigm that employs persistent homology on fMRI time series to quantify topological changes in functional connectivity patterns between MCI sub-types.\\
(ii) A comprehensive statistical analysis of topological features between HC and MCI sub-types is conducted to detect critical brain regions to examine distinct pattern linked with different stages of MCI across each homology dimension. 
To the best knowledge of authors, no other research has examined the efficacy of persistent homology of fMRI time series for differential diagnosis of MCI.

\begin{figure}[h]
\centering
\includegraphics[width=\textwidth]{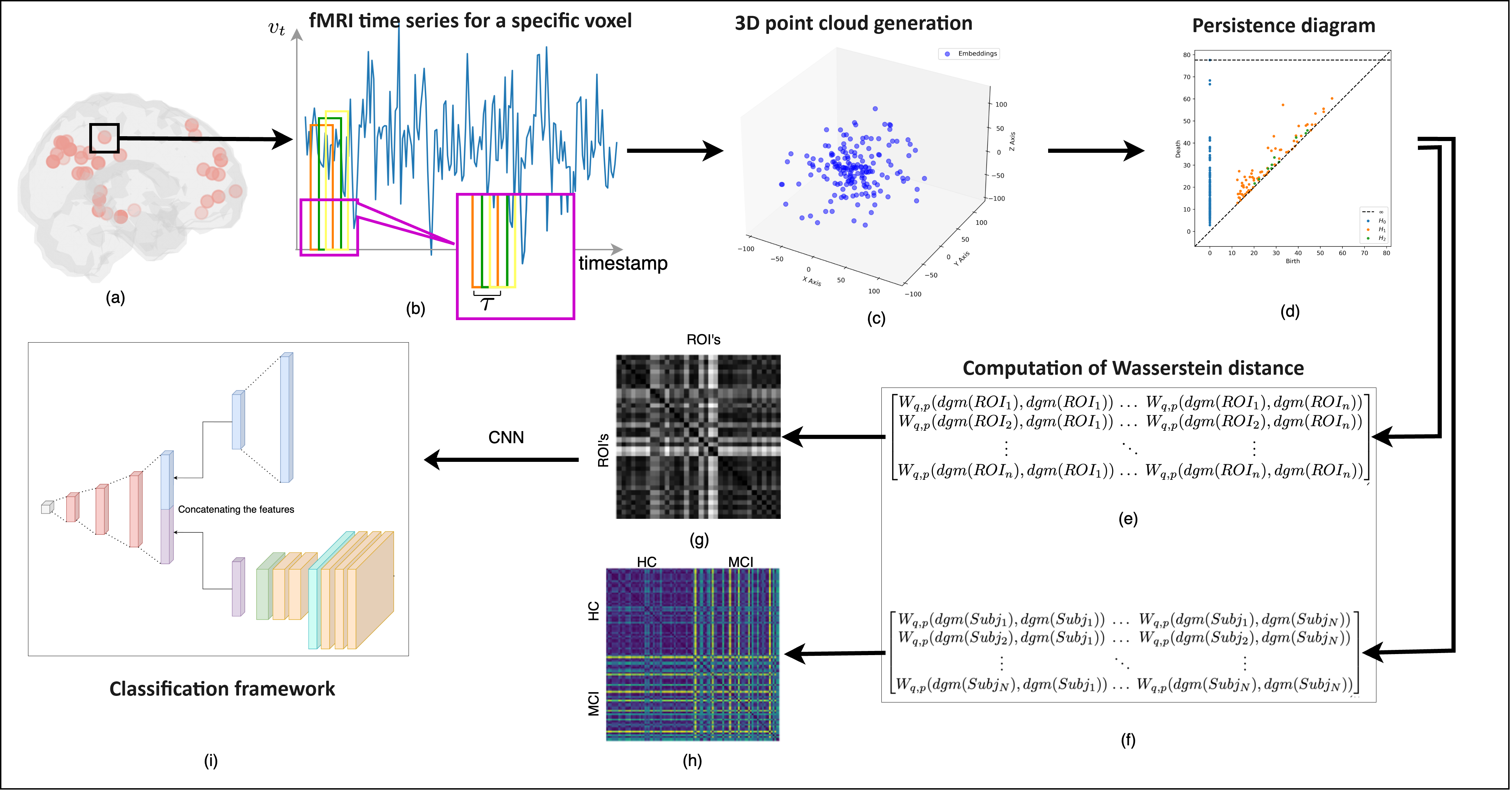}
\caption{ Block schematic of the proposed methodology
%Illustration of converting fMRI data pertaining to a ROI to 3D point cloud and using  persistent homology for comparing (i)interactions across different ROIs of a brain network, and (ii) ROI-specific topological dissimilarity across all HC and different stages of MCI subjects. \textbf{(a)} Brain ROI, \textbf{(b)} Time series extraction from each ROI and obtaining sliding window embedding with dimension $M=2$ and $\tau=1$ , \textbf{(c)} Projecting the embedding into $\mathbb{R}^3$ to form 3D point cloud, \textbf{(d)} Computation of $H_0,H_1, and H_2$ through persistent homology, \textbf{(e)} Matrix $PR$ depicting Wasserstein distance between all possible ROI pairs of a brain network, \textbf{(f)} Matrix $PS$ depicting Wasserstein distance across all study subjects of one specific ROI of a brain network. The corresponding sample image is shown in \textbf{(g)} that capture ROI interaction in a specific brain network and in \textbf{(h)} that capture the topological changes across all subjects of two classes (HC and MCI) for one specific ROI of a brain network, \textbf{(i)} Proposed CNN classification framework.
} 
\label{fig:blockdig}
\end{figure}

\section{Dataset description}

The study utilizes fMRI images from cohorts representing two distinct populations. The baseline study incorporates subjects from the publicly available Alzheimer’s Disease Neuroimaging Initiative (ADNI) dataset \cite{adni}, with a repetition time (TR) of 3000 ms and an echo time (TE) of 30 ms. This is complemented by our in-house TLSA (TATA Longitudinal Study for Aging) cohort, which has a TR of 3200 ms and a TE of 30 ms. The TLSA is an urban cohort investigation aimed at accumulating long-term data to discern risk and protective factors associated with dementia in India. Both cohorts feature images acquired in the sagittal acquisition plane with a 3D acquisition type. 
The subjects included from ADNI are: MCI ($N=50$), EMCI ($N=163$), LMCI ($N=141$) and HC ($N=179$). The efficacy of the proposed methodology is further verified on our in-house MCI ($N=50$) and HC ($N=50$) cohort. In our in-house MCI cohorts, individuals diagnosed with MCI met specific criteria, having clinical dementia rating or CDR value (the current gold standard for assessing the stages of patients diagnosed having dementia) of 0.5. 
All fMRI images underwent the same preprocessing pipeline, which included motion correction, adjustment for slice timing, normalization to the standard MNI space, and regression to account for nuisance variables. These preprocessing steps were performed using FSL (FMRIB Software Library) version 6.0.6 \cite{jenkinson2012fsl}.

%%%%%%%%%%%%%%%%%%%%%%%%%%%%%%%%%%%%%%%%%%%%%%%%%%%%%%%%%%%%%%%%
\section{Proposed Methodology}
The block diagram of the proposed methodology is shown in Figure~\ref{fig:blockdig}. Our method involves converting the 1D fMRI time series data into a 3D point cloud, from which persistent diagrams are obtained. We utilize the Wasserstein distance metric to compare all possible pairs of persistent diagrams, enabling quantification of topological alterations. These alterations are examined in two contexts: (i)\textit{across subjects for a given ROI} (ii)\textit{across ROIs for a given subject}. The in-house data will be made available to the research community after procedural formalities by the administration at the center.
\footnote{The codes of this analysis are available at https://github.com/blackpearl006/ICPR-2024}

\subsection{Extraction of time series}
The fMRI time series captures the temporal evolution of brain activity, providing a dynamic view of how different brain regions interact over time. The study carefully selects relevant brain regions from Dosenbach’s ROIs to extract fMRI time series for identifying meaningful patterns and differences between MCI and its sub-types.
The Dosenbach's ROIs\cite{dosenbach2010prediction} ($n=160$) are divided into six distinct brain networks: cerebellum ($\textit{n=18}$), cingulo-opercular ($\textit{n=32}$), default mode ($\textit{n=34}$), fronto-parietal ($\textit{n=21}$), occipital ($\textit{n=22}$), and sensorimotor ($\textit{n=33}$) networks, encompassing various interconnected brain regions. Each of these brain network is linked to specific cognitive, sensory, and motor functions and may show unique disruption patterns at various stages of MCI. Therefore, by analyzing all six networks, the study offers a comprehensive assessment of network-specific changes, which could serve as distinct biomarkers for different stages or types of cognitive impairment.
A representative of 5mm radius sphere drawn at each voxel location contributed as a time series \( \{ v_t, t = 1, 2, \ldots, N \} \).

\subsection{Point cloud construction from time series}
Creating an efficient point cloud representation from 1D time series data is a crucial step for computing persistent homology \cite{sw1,sw2}. 
In this step, the goal is to construct a richer feature space that enables the analysis of changes in the intrinsic topological properties of MCI. Each point in the time series is mapped to a vector in 3D space, in order to capture the complex temporal dynamics (Figure~\ref{fig:blockdig}.C).
For this, the study employs sliding window embedding ($\mathcal{SW}$) with embedding dimension $M = 2$ and time lag $\tau = 1$ in order to convert the fMRI time series $v_t$ into point clouds $\mathcal{S}$. The choice of length of the sliding window is set to $3$ to minimise noise interference and better interpretability.

The sliding window embedding of a function $f$ based at $t \in \mathbb{R}$ into $\mathbb{R}^{M+1}$ is represented as follows (Equation~\ref{eq:sliding}):
\begin{equation}
    \mathcal{SW}_{M,\tau} f: \mathbb{R} \rightarrow \mathbb{R}^{(M+1)}, \quad t \rightarrow \begin{bmatrix}
    f(t) \\
    f(t+\tau) \\
    f(t+2\tau) \\
    \end{bmatrix}
    \label{eq:sliding}
\end{equation}

Thus, selecting different values of $t$ results in a set of points, known as a sliding window point cloud for the function $f$. Multiple literature sources point to the efficiency of sliding window methods \cite{sliding1} in capturing dynamic functional connectivity in resting-state fMRI.
In this study, for embedding dimension $M = 2$, the point cloud of the fMRI time series is represented by Equation~\ref{eq:pointcloud}:

\begin{equation}
    \mathcal{S} = \{v_{i} : i = 1, \ldots, N, \ v_i \in \mathbb{R}^3\}
    \label{eq:pointcloud}
\end{equation}

This representation ensures each point $v_i$ in the point cloud $\mathcal{S}$ is a vector in $\mathbb{R}^3$, maintaining consistency with the spatial dimensions of the fMRI data.

\begin{figure}[h]
\centering
\includegraphics[width=0.5\textwidth]{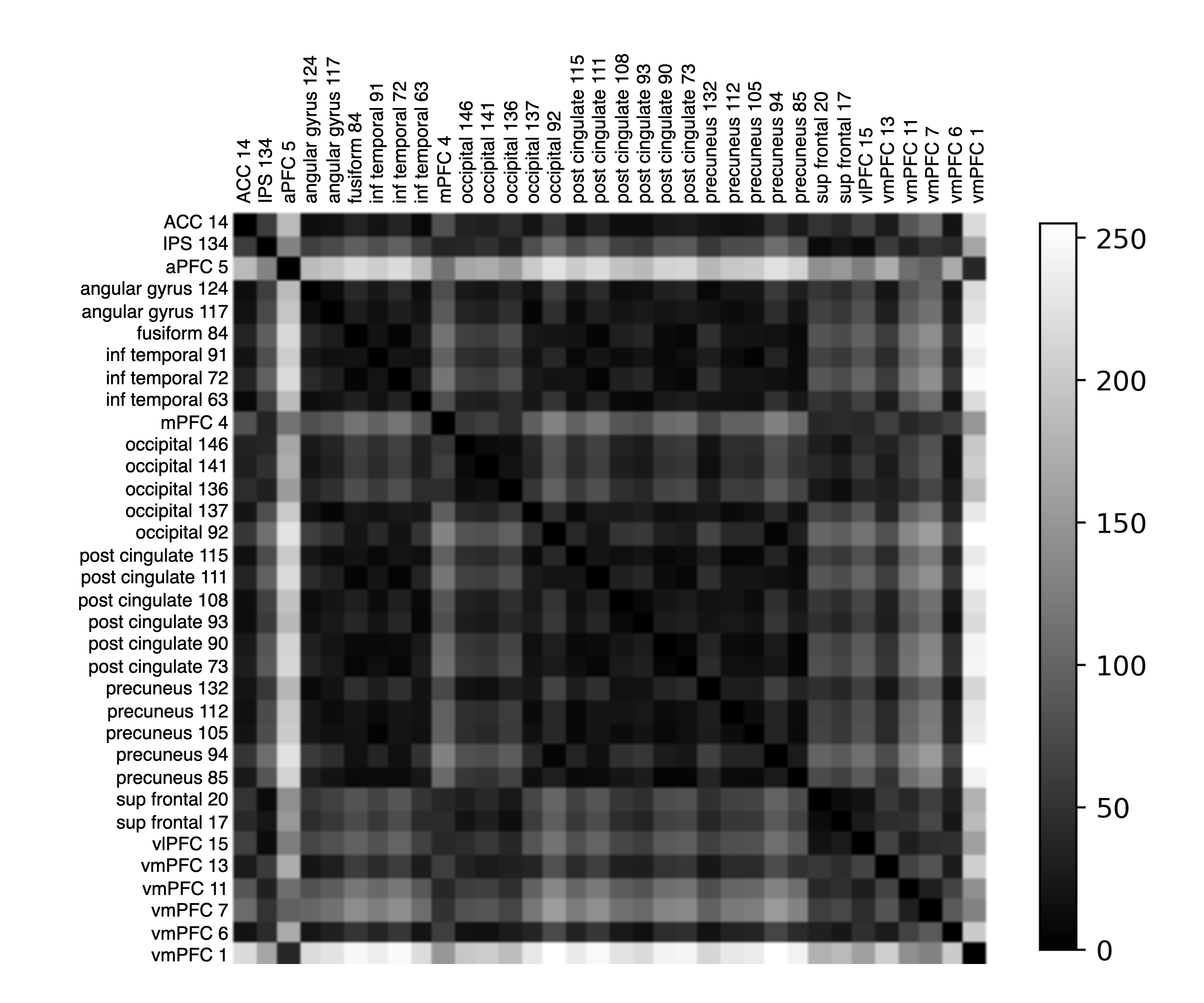}
\caption{Illustrating the Wasserstein distance as derived from the persistence diagram of fMRI time series for homology dimension-0, showing interactions among ROIs within the DMN for one representative subject. The 34 ROIs within the DMN exhibit a consistent spatial arrangement, wherein nearby brain regions are grouped together. The arrangement of these 34 ROIs remains consistent across all subjects. A low value of Wasserstein distance suggests synchronized neural activity and coordination between brain regions, while a high value of distance indicates distinct activity patterns and potential functional independence.}
\label{fig:dmn_roi_alignment}
\end{figure}

\subsection{Persistent homology}

This step involves computing persistent homology from the generated 3D point clouds to extract topological features of varying dimensions. This is accomplished by constructing a series of simplicial complexes using Vietoris-Rips filtration and calculating their homological features. These features capture the underlying topological structure in the data, identifying meaningful patterns and differences between MCI and its sub-types.
We exploit the information encoded in \textit{persistence diagram} to analyze the differences in topology of brain networks of individuals with MCI from HC.
Persistence diagram encodes the persistence features in data across the filtration parameter range as a collection of points in the two-dimensional Euclidean space $\mathbb{R}^2$. 
A common approach for constructing a filtration from a point cloud is through the Vietoris–Rips complex. This complex is generated from the point cloud by connecting any subset of points whose pairwise distances fall within a specified threshold, creating a simplex. Thus, filtration is a collection $\mathcal{F}= \{F_{\epsilon}\}_{\epsilon \geq 0} $ of spaces with $F_{\epsilon} \subset F_{\epsilon^{\prime}}$
continuous  $\forall \epsilon \leq \epsilon^{\prime}$. The $i^{th}$ persistence diagram of $\mathcal{F}$ is a multiset $dgm_{i}(\mathcal{F}) \subset \{(p,q) \in [0, \infty] \times [0, \infty] \mid 0 \leq p < q \} $ where each pair $(a,b) \in dgm_{i}(\mathcal{F})$ encodes a \textit{i}-dimensional topological feature, in other words Betti descriptors\footnote{In algebraic topology, the topological features of a space are represented as \emph{holes} or \emph{cycles} in various dimensions. The number of $k$-dimensional holes in a $d$-dimensional simplicial complex (with $k \leq d$) is denoted by the Betti number $\beta_k$ or $H_k$. Thus, $0$-dimensional holes $(\beta_0 or H_0)$ correspond to connected components, $1$-dimensional holes $(\beta_1 or H_1)$ represent tunnels (or loops) and $2$-dimensional holes $(\beta_2 or H_2)$ are voids.} associated with a simplicial complex that born at $F_{b}$ and dies at $F_{d}$. The quantity $(d-b)$ is the persistence of the feature, and typically measures significance across the filtration.
In our study, given a time series ($V_t$) the sliding window point cloud $\mathcal{SW}_{M,\tau} f$ is computed which is in a metric space $(X, M_{X})$. The Rips filtration $\mathcal{VR}(X,M_{X})$ is derived from the Vietoris–Rips complex $VR_{\epsilon}(X,M_{X})$, computed at each scale $\epsilon \geq 0$. The mathematical expression for computing Rips filtration is depicted in Equation~\ref{eq:vr}
\begin{multline}
    \mathcal{VR}(X,M_{X}):= \{VR_{\epsilon}(X,M_{X})\}_{\epsilon \geq 0}, where \\
    VR_{\epsilon}(X,M_{X}) := \{\{x_{0},..., x_{n}\} \in X \mid \max\limits_{0\leq i, j \leq n} M_{X}(x_{i},x_{j}) <  \epsilon, \textit{n} \in \mathbb{N}
\label{eq:vr}
\end{multline}
The birth-death pairs $(b,d)$ in the Rips persistence diagrams $dgm_{i}^{\mathcal{VR}} (X):= dgm_{i}\mathcal{VR}(X, M_{X})$ reveal the underlying topology of space $X$. The points $(b,d)$ in $dgm_{i}^{\mathcal{VR}} (X)$ with large persistence values $(d-b)$ suggest the most persistent topological features of the continuous space where $X$ is concentrated.

\par
The paper employs the Wasserstein distance to measure the dissimilarity between the persistence diagrams. This metric quantifies the differences in the topological features between subjects, providing a robust basis for distinguishing between different MCI sub-types. Thus, it enhances the ability to compare topological changes in brain activity. In this particular step, the paper contribution lies in analysing both inter-subject and inter-ROI Wasserstein distance measures to highlight the specific brain regions where changes are pronounced in MCI sub-types. 
In context of fMRI time series analysis using topological persistence diagram, the Wasserstein distance between fMRI time series of two brain regions captures the degree of similarity in their patterns of neural activity. A low Wasserstein distance indicates that the two fMRI time series exhibit similar patterns of neural activity over time. In other words, it suggests that the two brain regions are functionally synchronized and are likely engaged in coordinated activity. On the contrary, a high Wasserstein distance suggests that the fMRI time series of two brain regions have distinct patterns of neural activity, may be functionally dissociated or independent from each other. High Wasserstein distances may also indicate abnormalities in functional connectivity between the two brain regions, which could be indicative of any neurological or psychiatric disorders. Figure~\ref{fig:dmn_roi_alignment} illustrates the variability in Wasserstein distance among all ROI pairs for a single representative subject.
In computational topology, Bottleneck distance and Wasserstain distance are the two widely used measures for quantifying the dissimilarity between two persistence diagrams \cite{tdabook}. 
Suppose, $f_{1}$ and $f_{2}$ are two different filtrations and let $X=dgm_{p}(f_{1})$ and $Y=dgm_{p}(f_{2})$ denote the $p^{th}$ persistence diagrams corresponding to $f_{1}$ and $f_{2}$. The Wasserstein and Bottleneck distance metrices are used to quantify the dissimilarity between these two multisets $X$ and $Y$. Let $L_{\infty}(f_{1},f_{2})= \lVert f_{1}-f_{2} \rVert_{\infty} $ denote the supremum distance between $f_{1}$ and $f_{2}$, and $\eta$ denotes a bijection of $X \rightarrow Y $, then, the $q-$Wasserstein distance between two persistence diagrams $X$ and $Y$ is defined as 

\begin{equation}
W_{q,p}(X,Y) = \left[ \inf_{\eta:X\rightarrow Y} \sum_{x \in X} || x - \eta(x) ||_{\infty}^{q} \right]^{\frac{1}{q}}
\label{eq:wass}
\end{equation}

To compute the distance elements of $X$ and $Y$ one-to-one (bijection $\eta$) are matched. It is usually done in the following way: first for each pair of elements, $x \in X$ and $y = \eta(x) \in Y$, the difference between them (the cost function) is calculated using $|| x - \eta(x) ||_{\infty}$ that is basically $L_{\infty}$ norm. Adding up the $q^{th}$ degrees $||.||_{\infty}^{q}$, we get a notion of the difference between the whole multisets X and Y under the matching $\eta: X \rightarrow Y$. Taking the infimum over all possible bijections $\eta$, we get the difference between multisets $X$ and $Y$ under the best matching possible, effectively removing $\eta$ from further consideration. 
The bottleneck distance is the Wasserstein distance, with parameter  $q \rightarrow \infty$. Hence, one drawback of the bottleneck distance is its insensitivity to details of the bijection beyond the furthest pair of corresponding points. Due to this, the present study considers Wasserstein distance for quantification. \par
In this study, the persistence diagram is computed for two different scenarios: (i) \textit{ROI-specific-} across all ROIs of a particular subject, and (ii) \textit{Subject-specific-} across all considered subjects of HC and MCI for a particular ROI. Therefore, Wasserstein distance $W_{q,p}(X,Y)$ in Equation~\ref{eq:wass} is computed to measure the pairwise distance between all ROIs as well as between all subjects. This is described in the next sub-sections (Section~\ref{sec:roispecific} and Section~\ref{sec:subjspecific}).

%\vspace{-0.25cm}
\subsection{\textit{ROI-Specific} Inter-Subject Interactions}
\label{sec:roispecific}

%%%%%%%%%%%%%%%%%%%%%%%%%%%%%%%%%%%%%%%%%%%%%%%%%%%%
% \begin{equation}
%     \begin{bmatrix}
%     W(ROI_1,ROI_1) & \dots & W(ROI_1,ROI_n)) \\
%     W(ROI_1,ROI_2) & \dots & W(ROI_2,ROI_n) \\
%     \vdots & \ddots & \vdots \\
%     W(ROI_n,ROI_1) & \dots & W(ROI_n,ROI_n) \\
%     \end{bmatrix}_{n\times n}
%     \label{eq:pr_matrix}
% \end{equation}

%\begin{equation}
 %   \begin{bmatrix}
  %  W(Subj_1,Subj_1) & \dots & W(Subj_1,Subj_N)) \\
   % W(Subj_1,Subj_2) & \dots & W(Subj_2,Subj_N) \\
   % \vdots & \ddots & \vdots \\
    %W(Subj_N,Subj_1) & \dots & W(Subj_N,Subj_N) \\
    %\end{bmatrix}_{N\times N}
    %\label{eq:pr_matrix}
%\end{equation}
%%%%%%%%%%%%%%%%%%%%%%%%%%%%%%%%%%%%%%%%%%%%%%%%%%%%

ROI-specific analysis for all brain networks is conducted to understand the dissimilarity in topological patterns across all considered subjects for each of the three Betti descriptors ($H_0$, $H_1$, and $H_2$). This is represented in the pairwise-subject distance matrix ($PS$) of dimensions $N \times N$, where $N$ is the number of subjects. The mathematical representation of matrix $PS$ is shown in Equation~\ref{eq:ps}.
Each element $PS(i,j)$ indicates the Wasserstein distance between the persistence diagrams of subject $i$ and subject $j$ for a given ROI of a specific brain network. The sample images as obtained from $PS$ for one specific ROI of each brain network are shown in Figure~\ref{fig:eachROI}.
\begin{equation}
    PS=\begin{bmatrix}
    W_{q,p}(dgm(Subj_1), dgm(Subj_1)) & \dots & W_{q,p}(dgm(Subj_1), dgm(Subj_N)) \\
    W_{q,p}(dgm(Subj_2), dgm(Subj_1)) & \dots & W_{q,p}(dgm(Subj_2), dgm(Subj_N)) \\
    \vdots & \ddots & \vdots \\
    W_{q,p}(dgm(Subj_N), dgm(Subj_1)) & \dots & W_{q,p}(dgm(Subj_N), dgm(Subj_N)) \\
    \end{bmatrix}
    \label{eq:ps}
\end{equation}

\subsection{\textit{Subject-Specific} Inter-ROI Interactions}
\label{sec:subjspecific}

Here, for each of the three Betti descriptors ($H_0$, $H_1$, $H_2$), we compute the pairwise-ROI distance matrix ($PR$) with dimensions $n \times n$, where $n$ is the number of ROIs of a specific brain network. The mathematical representation of the matrix $PR$ is shown in Equation~\ref{eq:pr}. 
%The ROI alignment in the matrix $PR$  in itself contains group level information of ROI's, ROI's from the same brain region are closely placed, as shown in the Figure \ref{fig:dmn_roi_alignment} .
The Wasserstein distance in matrix $PR$ represents the interaction between different ROIs. Each element in $PR(i,j)$ indicates the distance between the persistence diagrams of ROI $i$ and ROI $j$. However, computing persistent homology and Wasserstein distance matrices ($PR$ and $PS$) are computationally very expensive, for which we utilised high-performance computing (HPC) resources, specifically an Intel(R) Xeon(R) Gold 6240 CPU @ 2.60GHz with dual CPUs and 192 GB of memory, were utilized.

\begin{equation}
   PR=\begin{bmatrix}
   W_{q,p}(dgm(ROI_1), dgm(ROI_1)) & \dots & W_{q,p}(dgm(ROI_1), dgm(ROI_n)) \\
    W_{q,p}(dgm(ROI_2), dgm(ROI_1)) & \dots & W_{q,p}(dgm(ROI_2), dgm(ROI_n)) \\
    \vdots & \ddots & \vdots \\
    W_{q,p}(dgm(ROI_n), dgm(ROI_1)) & \dots & W_{q,p}(dgm(ROI_n), dgm(ROI_n)) \\
    \end{bmatrix}
    \label{eq:pr}
\end{equation}

\subsection{Classification}
\begin{figure}[h]
\centering
\includegraphics[width=0.8\textwidth]{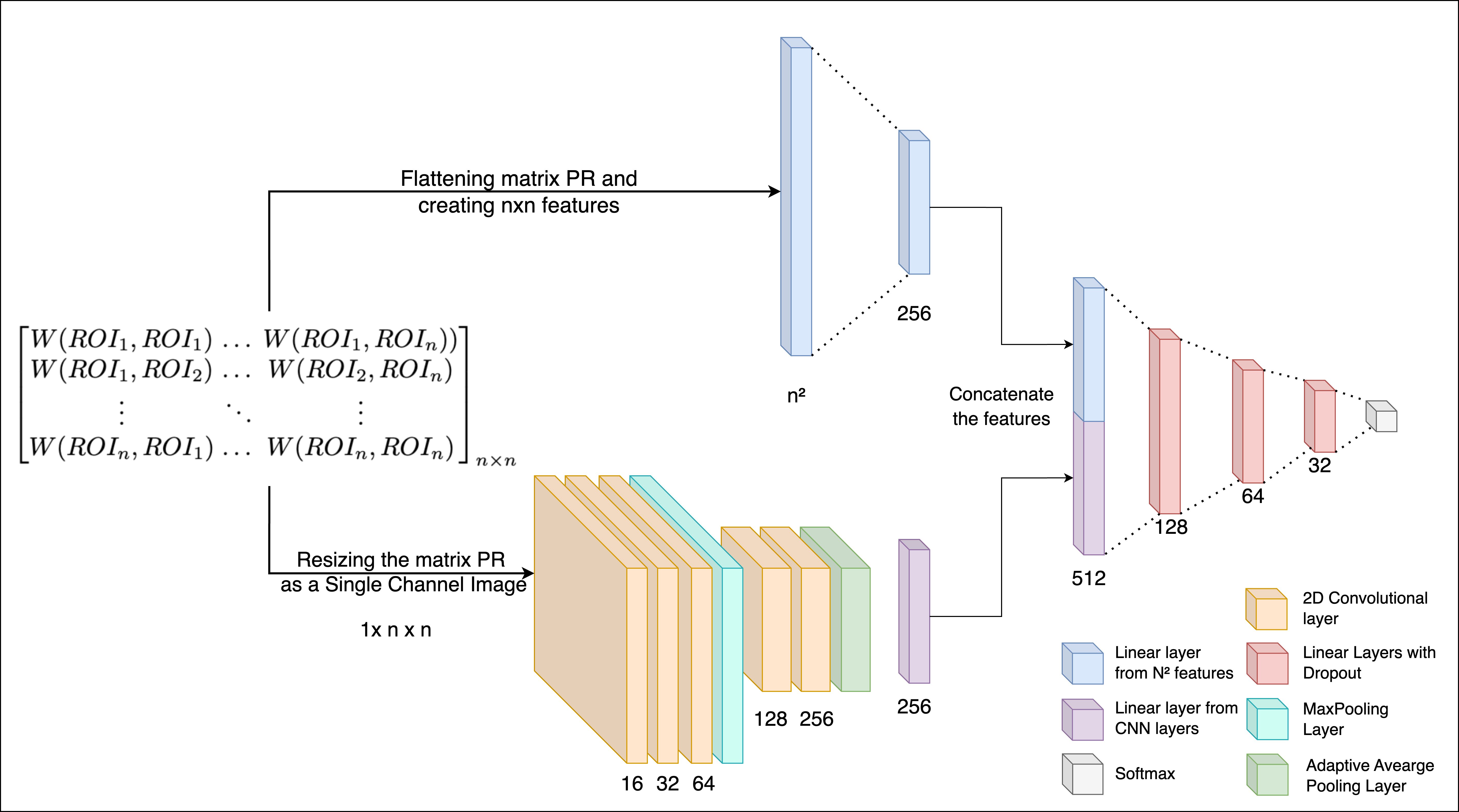}
\caption{The proposed deep learning architecture}
\label{fig:ourmodel}
\end{figure}

The study integrates the 1D and 2D features from Wasserstein distances that captures the inter-ROI interaction for each subject into a classification framework using conventional CNN. The classification results demonstrate the effectiveness of the proposed method in distinguishing between different stages of MCI, showcasing the practical applicability of persistent homology in medical diagnostics. 
To perform classification, the subject-specific $PR$ matrix ($n\times n$) is utilized. The proposed CNN architecture used for classification purpose is depicted in Figure \ref{fig:ourmodel}. This model integrates 1D features extracted from each ROI pair of the $PR$ matrix with the 2D CNN features. 
Thus, the proposed classification model includes two steps. First, the Wasserstein distance matrix is flattened to create 1D features, focusing on pairwise relationships. In the second step, 2D features are extracted from the distance matrix through CNN layers to capture local patterns and spatial hierarchies (Figure~\ref{fig:dmn_roi_alignment}). The features from the flattened matrix and the CNN layers are concatenated, creating a unified feature vector that combines information from both linear and convolutional layers. The concatenated features are then passed through several dense layers with dropout for regularization, reducing the risk of overfitting. Combining 1D and 2D features allows the model to create a richer and more diverse feature space that can capture different aspects of the data. 1D features can represent sequential or linear relationships, while 2D features can capture spatial or topological relationships. Hence, integrating both 1D and 2D features offer several advantages including a comprehensive and richer feature space for classification, enhanced learning from different perspectives, helps in mitigating the impact of noise and artifacts, and the ability to capture both local and global patterns. 
For the 2D features, our model comprises three CNN layers with 16, 32, and 64 filters respectively. This is followed by a max-pooling layer and another convolutional block with two CNN layers having 128 and 256 filters. We use kernel size of 3 for all CNN layers. Subsequently, a global average pooling layer condenses each feature map into a single value, forming a linear feature vector. ReLU activation functions are employed throughout the model.
Simultaneously, the 1D features of the $PR$ matrix ($n^2 \times 1$), are processed through a linear layer, reducing them to 256 features. The resulting 256-dimensional feature vector from this linear layer and the 256-dimensional feature vector from the 2D CNN are concatenated and fed as an input to a series of fully connected layers of size 128, 64, and 32, each incorporating dropout set to 0.2. A softmax activation function is applied at the final layer for classification.
Each Betti descriptor ($H_{0}$, $H_{1}$, and $H_{2}$) from every brain network is independently analyzed. The model is trained over 100 epochs using the Adam optimizer with a learning rate of 0.001 and a train-test split of 80-20. Cross-Entropy loss is employed for training. The model is trained to classify the following scenarios: (i) MCI versus HC (for both ADNI and our in-house dataset), (ii) EMCI versus HC (ADNI), (iii) LMCI versus HC (ADNI), and (iv) EMCI versus LMCI (ADNI). The classification part of this experiment is conducted using Kaggle notebooks with $2\times 16$GB NVIDIA Tesla T4 GPU and the PyTorch deep learning framework.
%%%%%%%%%%%%%%%%%%%%%%%%%%%%%%%%%%%%%%%%%%%%%%%%%%%%%%%%%%%%%%%%
%\vspace{-0.5cm}
\section{Results and Discussion}

The current study examines alterations in brain network topology between HC and MCI sub-types using persistent homology of fMRI time series. Numerous studies have been conducted over past decades to classify MCI from HC, with several groups reporting fair classification performance. However the novelty in the proposed method lies in incorporation of an innovative methodology that combines sliding window embedding of fMRI time series with a tool originating from the emerging field of computational topology. Furthermore, an extensive statistical analysis on topological features comparing HC and MCI sub-types is carried out, aiming to identify statistically significant ROI-pairs in each functional brain network for investigating unique neurobiological patterns associated with various stages of MCI across different homology dimensions.  \par
\begin{figure}[h]
\centering
\includegraphics[width=\textwidth]{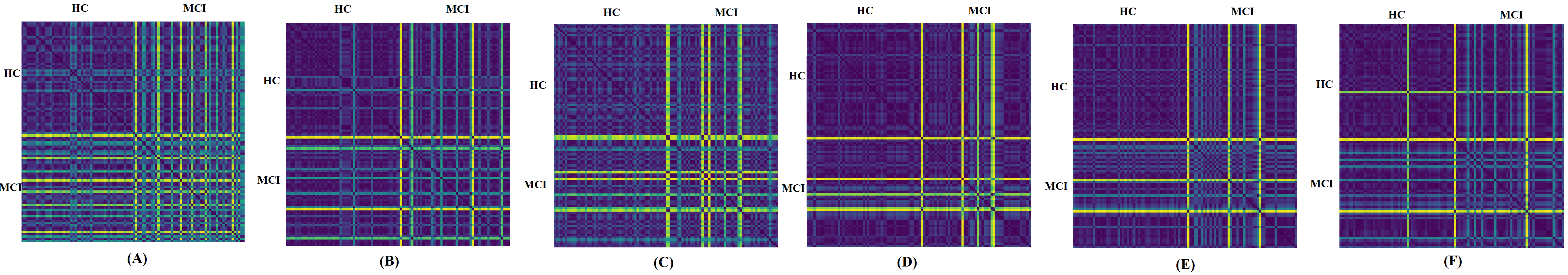}
\caption{ROI-specific Wasserstein distance (dim-0) across all subjects of HC (\textit{n=50}) and MCI (\textit{n=50}) from each considered brain network- showing the visible difference in pattern of Wasserstein distance between HC and MCI subjects. These ROIs are (A) Post cingulate 108 (DMN), (B) inf cerebellum 121 (CB), (C) IPL 96 (FP), (D) Occipital1 106 (OP), (E) Post cingulate 80 (CO) and (F) Pre-SMA 41 (SM).} 
\label{fig:eachROI}
\end{figure}

 
%The generation of 3D vector point cloud from fMRI time series corresponding to each ROI of a specific brain network is shown in Figure~\ref{fig:blockdig}(c). The choice of length of the sliding window is set to $3$ to minimise noise interference and better interpretability.
Persistence homology is computed for dimension-0, 1, and 2 on each point cloud using Vietoris-Rips filtration. The computation of persistent homology is performed using Ripser software \cite{ripsersoftware}. The filtrations provide a basis for computing the persistent topological features that exist within the point cloud. The persistence of topological features is then encoded in persistence diagram. For three different homology dimensions $H_0$, $H_1$ and $H_2$, the derived persistence diagram for a point cloud is shown in Figure~\ref{fig:blockdig}(d). As described in methodology section, in the persistence diagram, every point corresponds to a specific topological feature. The magnitude of the difference between the ``birth" ($b$) and ``death" ($d$) values indicates the life-span or persistence of the topological descriptors. The Wasserstein distance metric is used to compute the dissimilarity between two persistence diagrams. Figure~\ref{fig:dmn_roi_alignment} and Figure~\ref{fig:eachROI} capture this dissimilarity across all ROIs and across all subjects, respectively.

\begin{figure}[h]
\centering
\includegraphics[width=\textwidth]{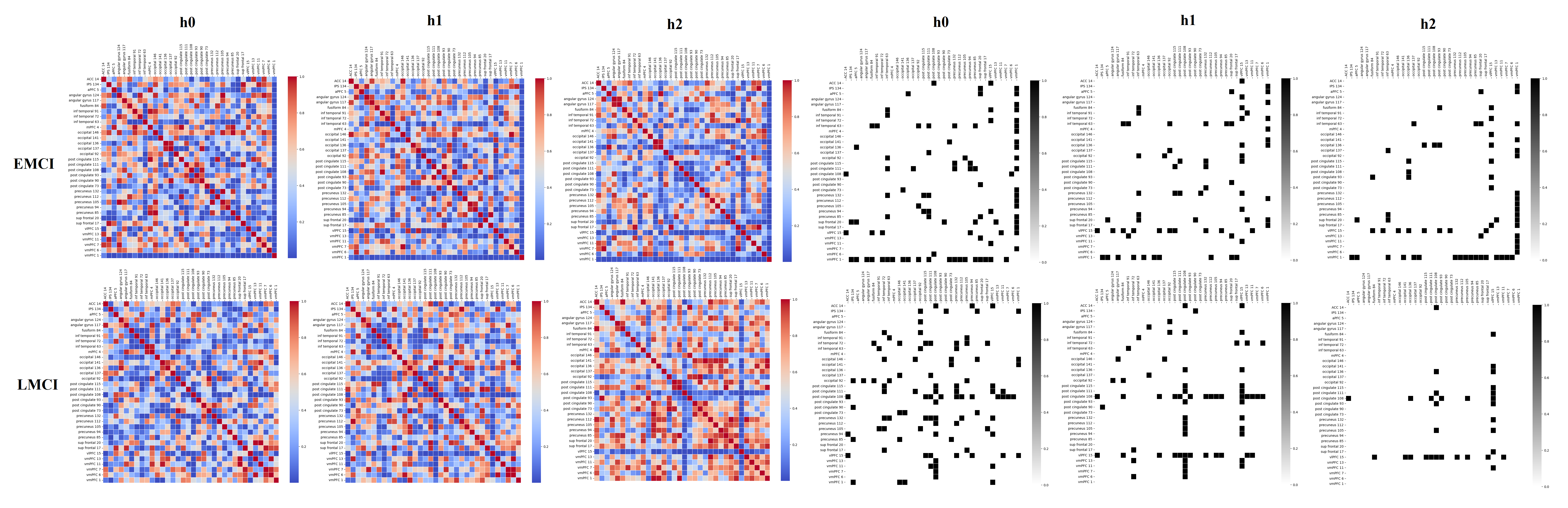}
\caption{Visualization of P-plot at 99\% C.I for Default mode network for EMCI (\textit{top row}) and LMCI \textit{bottom row}. The \textit{column 1, 2 and 3} show the P-plot for homology dimension 0, 1 and 2, respectively. \textit{Column 4, 5, and 6} highlight the ROI-pair that showed significant differences $(p<0.01)$ in topology across all subjects of between (i) HC and EMCI, (ii) HC and LMCI. The visualization clearly depicts significant ROIs which are seen to be more concentrated in fewer brain regions as homology dimension increases in case of LMCI as compared to EMCI.}
\label{fig:pval_dmn}
\end{figure}

\begin{table}[ht]
\caption{The classification accuracy as obtained across six distinct brain networks.
}
\centering
\begin{tabular}{|l|p{2cm}|p{1cm}|p{1cm}|p{1cm}|p{1cm}|p{1cm}|p{1cm}|p{1cm}|}
\hline
\textbf{Comparison}  & \textbf{Dataset} & \textbf{Dim.}& \textbf{DMN} & \textbf{FP} & \textbf{OP} & \textbf{SM} & \textbf{CO} & \textbf{CB} \\
\hline
HC vs MCI &ADNI& $H_{0}$ & 70.0  & 90.0 & 85.0 & 85.0 & \textbf{95.0} & 85.0 \\
\cline{3-9}
 & & $H_{1}$ & 65.0  &90.0 & \textbf{90.0} & 75.0 & 75.0 & 75.0 \\
\cline{3-9}
 & & $H_{2}$ & 75.0  & 80.0 & 85.0 & 85.0 & 75.0 & 80.0 \\
\hline
HC vs MCI &In-house & $H_{0}$ &82.4  & 71.4 & 57.8 & \textbf{85.0} & 70.6 & 63.0 \\
\cline{3-9}
 &TLSA & $H_{1}$ & 76.5  & 52.9 & 68.4& 72.2 & 52.9 & 63.2 \\
\cline{3-9}
 & & $H_{2}$ & 70.6  & 64.7 & 73.8 & 55.0 & 76.5 & 57.9 \\
\hline
HC vs EMCI & ADNI & $H_{0}$ & \textbf{76.5} & 60.3 & 60.3 & 58.8 & 66.2 & 64.7 \\
\cline{3-9}
 & & $H_{1}$ & 61.8 & 69.1 & 75 & 57.4 & 52.9 & 60.3 \\
\cline{3-9}
 & & $H_{2}$ & 66.2 & 50 & 69.1 & 55.9 & 61.8 & 55.9 \\
\hline
HC vs LMCI & ADNI & $H_{0}$ & \textbf{91.1} & 76.8 & 84.1 & 80.1 & 74.6 & 71.4 \\
\cline{3-9}
 & & $H_{1}$ & 85.7 & 55.6 & 79.3 & 74.6 & 73 & 61.9 \\
\cline{3-9}
 & & $H_{2}$ & 63.5 & 65.1 & 58.7 & 66.6 & 53.9 & 60.3 \\
\hline
EMCI vs LMCI &ADNI & $H_{0}$ & \textbf{80.0} & 65.0 & 71.7 & 76.7 & \textbf{80.0} & 71.7 \\
\cline{3-9}
 & & $H_{1}$ & 75.0 & 66.7 & 68.3 & 70.0 & 61.6 & 68.3 \\
\cline{3-9}
 & & $H_{2}$ & 56.7 & 53.3 & 53.3 & 56.7 & 56.6 & 60.0 \\
\hline

%\cline{3-9}
%\hline
%HC vs MCI &Train:ADNI & h0 & 56.3 & 57.0 & 61.2 & 50.6 & 57.5 & 50.0 \\
%\cline{3-9}
% & Test:TLSA & h1 & 52.9 & 54.1 & 53.1 & 50.5 & 51.7 & 51.0 \\
%\cline{3-9}
% & & h2 & 52.9 & 50.0 & 64.3 & 52.8 & 60.9 & 52.0 \\
%\hline

%HC vs MCI &Train:TLSA & h0 & 52.0 & 50.0 & 53.0 & 52.0 & 50.0 & 50.0 \\
%\cline{3-9}
% & Test:ADNI & h1 & 52.0 & 54.0 & 51.0 & 52.0 & 64.7 & 54.0 \\
%\cline{3-9}
% & & h2 & 55.0 & 54.0 & 50.0 & 51.0 & 54.0 & 54.4 \\
%\hline
\end{tabular}
\label{tab:tablenew}
\end{table}

%\begin{table}[ht]
%\caption{Logistic Regression classification accuracy for ADNI HC vs MCI, including Densenet classification accuracy on ADNI, where the input is a 3 channel image with each channel corresponding to a betti descriptor's Wasserstein disatnce matrix}
%\centering
%\begin{tabular}{|l|l|l|l|l|}
%\hline
%\multirow{2}{*}{Brain Network} & \multirow{2}{*}{ Densenet } & \multicolumn{3}{c|}{ %Logistic Regression } \\ \cline{3-5} 
 %                             &                           & h0      & h1      & h2      \\ \hline
%DMN                           & 80.0\%                    & 78.27   & 72.42   & 57.26   \\ \hline
%CO                            & 90.0\%                    & 75.32   & 71.05   & 56.27   \\ \hline
%FP                            & 96.0\%                    & 64.12   & 63.16   & 58.86   \\ \hline
%OP                            & 74.4\%                    & 74.01   & 67.1    & 59.19   \\ \hline
%CB                            & 85.0\%                    & 70.33   & 69.97   & 56.43   \\ \hline
%SM                            & 80.0\%                    & 75.99   & 70.73   & 61.2    \\ \hline
%\end{tabular}
%\label{tab:brain-network-accuracy}
%\end{table}

\begin{table}[ht]
\caption{Comparison of peak classification accuracy of the proposed CNN architecture with the conventional Densenet-121 and also with random forest ensemble classifier to check the efficacy of the proposed model in classifying HC and MCI.
}
\centering
\begin{tabular}{|l|l|l|l|l||l|l|}
\hline
 
Brain &\multicolumn{2}{c|}{Densenet-121}  &\multicolumn{2}{c||}{Ensemble Classifier} &\multicolumn{2}{c|}{\textbf{Proposed Model}} \\ 
\cline{2-7}
Network & ADNI & In-house  & ADNI & In-house & ADNI & In-house \\
 & &TLSA & &TLSA & &TLSA\\
\hline
DMN      & 80.0\%   & 82.0\%    & 67.0\%  & 72.48\%   &\textbf{91.1\%}  &\textbf{82.4\% } \\ \hline
FP       & 89.6\%   &71.4\%    & 83.0\%   & 61.67\%  &\textbf{90.0\%}  &\textbf{71.4\%}  \\\hline
OP       & 74.4\%   &72.9\%    & 84.0\%   & 70.67\%  &\textbf{90.0\%}   &\textbf{73.8\%}  \\ \hline
SM       & 80.0\%   &72.2\%    & 80.0\% & 69.3\%  &\textbf{85.0\%}   &\textbf{85.0\%}  \\ \hline
CO       & 90.0\%   &75.0\%    & 86.0\%  & 68.19\% &\textbf{95.0\%}   &\textbf{76.5\% } \\  \hline
CB       & 85.0\%    &62.6\%   & 82.0\%  & 63.44\%    &\textbf{85.0\%}   &\textbf{63.2\%}  \\ 
\hline

\end{tabular}
\label{tab:tablenew1}
\end{table}

\begin{table}[h]
\caption{Comparative analysis with recent state-of-the-art techniques that used fMRI data to differentiate disease sub-types as well as to distinguish MCI and its sub-types.}
\centering
    \begin{tabular}{|l|l|l|l|l|}
        \hline
        \textbf{Reference} & \textbf{Year} & \textbf{Modality (dataset)} & \textbf{Subjects} & \textbf{Accuracy} \\
        \hline
        \cite{refcomp1new} & 2022 & fMRI (local) & HC Vs. MCI & 65.14\% \\
        \hline
        \cite{refcomp2new} & 2023 & fMRI (ADNI) & HC Vs. MCI & 87\% \\
        \hline
        \cite{refcomp3} & 2018 & fMRI (ADNI) & HC Vs. MCI & 82.6\% \\
         &  &  & EMCI Vs. LMCI &74.3\% \\
        \hline
        \cite{refcomp4} & 2021 & fMRI (ADNI) & HC Vs. MCI & 82.8\% \\
        \hline
        \cite{refcomp5} & 2020 & fMRI (ADNI) &HC Vs. MCI & 80\% \\
        \hline
        \cite{our1} & 2024 & fMRI (ADNI) & HC Vs. MCI & 89.47\% \\
        \hline
        \cite{refcompicpr1} & 2018 & fMRI (ADNI) & HC Vs. EMCI & 74.23\% \\
        \hline
        \cite{refcompicpr2} & 2020 & fMRI (ADNI) & HC Vs. EMCI & 76.07\% \\
        \hline
        \cite{refcompicpr3} & 2021 & fMRI (ADNI) & HC Vs. EMCI & 74.42\% \\
        \hline
        \cite{refcompicpr5} & 2021 & fMRI (ADNI) & HC Vs. LMCI & 87.2\% \\
        \hline
        \cite{refcompicpr4} & 2021 & fMRI (ADNI) & EMCI Vs. LMCI & 79.36\% \\
        \hline
        \hline
        \textbf{Proposed} & 2024 & fMRI (ADNI) & HC Vs. MCI & \textbf{95\% ($H_{0}$)}\\
        \cline{3-5}
         \textbf{method} &  &fMRI (in-house TLSA)  & HC Vs. MCI & \textbf{85\% ($H_{0}$)}  \\
        \cline{3-5}
         & &fMRI (ADNI) &HC Vs. EMCI & \textbf{76.5\% ($H_{0}$)}  \\
         \cline{4-5}
         & & &HC Vs. LMCI & \textbf{91.1\% ($H_{0}$)}  \\
         \cline{4-5}
          & & &EMCI Vs. LMCI & \textbf{80\% ($H_{0}$)}  \\
         \hline          
    \end{tabular}
    \label{tab:tablecomp}
\end{table}

From persistence diagram both inter-subject ($PS$) and inter ROI ($PR$) Wasserstein distance is computed for each homology dimension as shown in Figure~\ref{fig:blockdig}.(g) and Figure~\ref{fig:blockdig}.(h). The Wilcoxon rank-sum test is conducted at 99\% C.I on PR matrix to identify key ROI-pairs within brain and their associated patterns in order to distingusish EMCI and LMCI. As discussed previously, since the Wasserstein distance between fMRI time series of two brain regions captures the degree of topological similarity in their neural activation pattern, our study hypothesis seeks to find ROI-pairs for which the inter-ROI Wasserstein distance between (i) HC Vs. EMCI, (ii) HC Vs. LMCI groups is statistically significant $(p<0.01)$. Identification of such ROI-pairs in EMCI and LMCI may aid to uncover distinct neurobiological signatures associated with different stages of MCI.
The visualization of the p-value plot which we refer as \textit{P-plot}, as obtained for each ROI pair of DMN is shown in Figure~\ref{fig:pval_dmn}. The P-plot visualization shows clear disparity in ROI pairs between EMCI and LMCI for each homology dimension-0 (\textit{column-4}), 1 (\textit{column-5}) and 2 (\textit{column-6}). It is seen that significant ROI pairs tend to concentrate within fewer regions of DMN in the case of LMCI as compared to EMCI. This pattern is seen to be consistent for all six brain networks and pronounced particularly for homology dimension 2, followed by dimension 1. For example, as seen from Figure~\ref{fig:pval_dmn}, in LMCI, significant dissimilarities in brain activity with other ROIs are primarily concentrated in two regions of DMN: post-cingulate-108 (PC108) and Ventrolateral Prefrontal Cortex (vlPFC). Conversely, in EMCI, significant dissimilarities in brain activity with other ROIs are extended to additional regions such as vmPFC, occipital, and inf temporal, in addition to vlPFC and PC108, for homology dimension-2. For dimension-1 also this spread in pattern across DMN regions is more noticeable for EMCI than LMCI cases. Post cingulate cortex plays a central role in various cognitive functions, including memory retrieval, attention, and self-referential processing. The vlPFC is portion of the prefrontal cortex which is located on the inferior frontal gyrus, involved in higher-order cognitive functions and executive control such as decision-making, response inhibition, working memory, and goal-directed behavior. Thus, the significant topological dissimilarities in activation pattern the post cingulate regions and vlPFC with several other brain regions of DMN in LMCI as compared to EMCI likely reflects the progressive neurodegenerative changes associated with advanced stages of cognitive impairment.
\begin{figure}[h]
\centering
\includegraphics[width=0.7\textwidth]{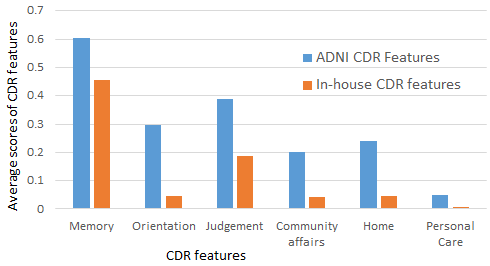}
\caption{Illustrating the variation in six standard clinical dementia rating features between two distinct population: (i) ADNI cohort and (ii) in-house TLSA cohort.} 
\label{fig:CDRscores}
\end{figure}
These findings shed light on differences in the evolving patterns of neural activity and functional connectivity within specific brain regions, offering potential biomarkers for differential diagnosis. Furthermore, the significant dissimilarities in Wasserstein distance within fewer regions particularly in the PCC and vlPFC of DMN, in LMCI compared to EMCI implies a progression towards more localized disruptions in brain activity as cognitive impairment advances. These insights deepen our understanding of the topological changes occurring in neural networks as cognitive impairment progresses, offering new directions for tailored diagnostic and therapeutic interventions.\par

The performance of the proposed model in classifying MCI and its sub-types is tabulated in Table~\ref{tab:tablenew}. The classification using the proposed CNN model yields the highest accuracy of $95\%$ and $85\%$ for ADNI and in-house TLSA MCI cohort, respectively. In case of MCI sub-types classification from HC, the accuracy is decreased to 76.5\% to classify EMCI from HC. However, in classifying LMCI from HC, highest accuracy of 91.1\% is obtained. While distinguishing disease sub-types EMCI Vs. LMCI, 80\% accuracy is obtained. 
The variability in classification accuracy within the DMN can be attributed to the degree of cognitive impairment that influence the model's ability to distinguish between MCI sub-types.
While comparing the peak accuracy across all networks, it is found that 0-dimensional topological features ($H_{0}$) perform the best. This is because $H_{0}$ captures the most fundamental topological aspect of data which is the number of connected components. This translates to identifying how functional connectivity among different regions in a specific brain network evolve over time. In neurodegenerative disease like MCI, initial changes often manifest as alterations in basic connectivity pattern rather than more complex topological structures like loops ($H_{1}$) or voids ($H_{2}$). This makes $H_{0}$ a reliable measure for detecting such changes, leading to better classification performance. Moreover, $H_{0}$ features are simpler and less prone to noise as compared to higher-dimensional features. These complex features might introduce variability that does not contribute meaningfully to classification performance. 
To validate the efficacy of the proposed CNN model, its results are compared with those of the classical DenseNet-121 architecture and the random forest, which has shown superior performance among other random forest ensemble classifiers. It has been observed that the deep learning model generally outperforms traditional machine learning classifiers. On the ADNI dataset, the proposed CNN model surpasses both the DenseNet-121 and the ensemble classifier. For the TLSA dataset, the performance of the proposed CNN model is comparable to that of the DenseNet-121 model across all brain networks. This is shown in Table~\ref{tab:tablenew1}. 
However, the notable differences in classification accuracy are observed between the two distinct population. This can arise due to several factors. Population from different regions or ethnicity may have distinct demographic characteristics, genetic backgrounds, lifestyle factors, cultural practices, socioeconomic status, education levels, environmental exposures and prevalence rates of certain diseases. These differences can contribute to variations in brain structure, function, and connectivity patterns, affecting the results of fMRI analyses.  Figure~\ref{fig:CDRscores} shows the clear differences in six standard CDR features as obtained for the two distinct population. Moreover, differences in data acquisition protocols and imaging parameters may result in variations in the functional connectivity patterns between datasets of distinct population. 
Hence, the present study reports the classification performance separately on two distinct populations to ensure that the data is as comparable as possible and not confounded by site differences. This approach enhances the generalization ability and reliability of the proposed CNN methodology.
It is seen that the proposed methodology outperforms the recent state-of-the art techniques that utilized fMRI to study MCI. The table comparing the proposed method with SOTA is shown in Table~\ref{tab:tablecomp}.
The clinical relevance of the study lies in its ability to localize critical regions in the brain network whose activity patterns and connectivity are altered across different stages of MCI. By identifying these key regions in brain network and their associated topological patterns (Figure~\ref{fig:pval_dmn}), the study provides valuable insights into the progression of cognitive impairment and the underlying neurobiological changes. The findings from the current study suggest the potential utility of employing persistent homology for differential diagnosis of MCI.
Nevertheless, further research is required in order to validate these findings across diverse cohorts with more samples for conclusive inferences.

\section{Acknowledgement} We thank the Director, Dr. K.V.S. Hari and the administration of Centre for Brain Research, IISc, for the support provided throughout the study.

\bibliographystyle{splncs04}
\bibliography{ref}

\begin{thebibliography}{10}
\providecommand{\url}[1]{\texttt{#1}}
\providecommand{\urlprefix}{URL }
\providecommand{\doi}[1]{https://doi.org/#1}

\bibitem{ninad}
Aithal, N., Pradeep, C.S., Sinha, N.: Mci detection using fmri time series embeddings of recurrence plots. In: 2024 IEEE International Symposium on Biomedical Imaging (ISBI). pp.~1--4 (2024). \doi{10.1109/ISBI56570.2024.10635716}

\bibitem{ammu1}
Ammu, R., Sinha, N.: Analysis of mild cognitive impairment utilizing covariance matrices of brain regions. In: 2023 IEEE 33rd International Workshop on Machine Learning for Signal Processing (MLSP). pp.~1--6. IEEE (2023)

\bibitem{ripsersoftware}
Bauer, U.: Ripser: efficient computation of vietoris--rips persistence barcodes. Journal of Applied and Computational Topology  \textbf{5}(3),  391--423 (2021)

\bibitem{our1}
Bhattacharya, D., , Sinha, N.e.: Multi-scale fmri time series analysis for understanding neurodegeneration in mci. arXiv preprint arXiv:2402.02811  (2024)

\bibitem{boldfmrits}
Bhattacharya, D., Sinha, N., Chattopadhyay, A., et~al.: Image complexity based fmri-bold visual network categorization across visual datasets using topological descriptors and deep-hybrid learning. arXiv preprint arXiv:2311.08417  (2023)

\bibitem{refcomp5}
Bi, X.a., Hu, X.e.: Multimodal data analysis of alzheimer's disease based on clustering evolutionary random forest. IEEE Journal of Biomedical and Health Informatics  \textbf{24}(10),  2973--2983 (2020)

\bibitem{refcomp2new}
Bolla, G., Berente, D.B.e.: Comparison of the diagnostic accuracy of resting-state fmri driven machine learning algorithms in the detection of mild cognitive impairment. Scientific Reports  \textbf{13}(1),  22285 (2023)

\bibitem{2021-Caputi-Epilepsy}
Caputi, L., Pidnebesna, A., Hlinka, J.: Promises and pitfalls of topological data analysis for brain connectivity analysis. NeuroImage  \textbf{238},  118245 (2021)

\bibitem{2022-Das-TDA-Brain-Networks}
Das, S., Anand, D.V., Chung, M.K.: Topological data analysis of human brain networks through order statistics. Plos one  \textbf{18}(3),  e0276419 (2023)

\bibitem{dosenbach2010prediction}
Dosenbach, N.U., Nardos, B.e.: Prediction of individual brain maturity using fmri. Science  \textbf{329}(5997),  1358--1361 (2010)

\bibitem{2016-Dunaeva-Endoscopy-Analysis}
Dunaeva, O., Edelsbrunner, H.e.: The classification of endoscopy images with persistent homology. Pattern Recognition Letters  \textbf{83},  13--22 (2016)

\bibitem{tdabook}
Edelsbrunner, H., Harer, J.L.: Computational topology: an introduction. American Mathematical Society (2022)

\bibitem{new1}
Farias, S.T., Mungas, D., Reed, B.R., Harvey, D., DeCarli, C.: Progression of mild cognitive impairment to dementia in clinic-vs community-based cohorts. Archives of neurology  \textbf{66}(9),  1151--1157 (2009)

\bibitem{sw2}
Gakhar, H., Perea, J.A.: Sliding window persistence of quasiperiodic functions. Journal of Applied and Computational Topology pp. 1--38 (2023)

\bibitem{sliding1}
Hindriks, R., Adhikari, M.H.e.: Can sliding-window correlations reveal dynamic functional connectivity in resting-state fmri? Neuroimage  \textbf{127},  242--256 (2016)

\bibitem{refcomp1new}
Hu, M., Yu, Y.e.: Classification and interpretability of mild cognitive impairment based on resting-state functional magnetic resonance and ensemble learning. Computational intelligence and neuroscience  \textbf{2022} (2022)

\bibitem{adni}
Jack, J., Clifford, R.e.: The alzheimer's disease neuroimaging initiative (adni): Mri methods. Journal of Magnetic Resonance Imaging  \textbf{27}(4),  685--691 (2008)

\bibitem{janoutova2015mild}
Janoutov{\'a}, J., Ser{\`y}, O.e.: Is mild cognitive impairment a precursor of alzheimer's disease? short review. Central European journal of public health  \textbf{23}(4), ~365 (2015)

\bibitem{jenkinson2012fsl}
Jenkinson, M., Beckmann, C.F.e.: Fsl. Neuroimage  \textbf{62}(2),  782--790 (2012)

\bibitem{refcomp3}
Jie, B., Liu, M.e.: Sub-network kernels for measuring similarity of brain connectivity networks in disease diagnosis. IEEE Transactions on Image Processing  \textbf{27}(5),  2340--2353 (2018)

\bibitem{refcompicpr2}
Kam, T.E., Zhang, H., Jiao, Z.e.a.: Deep learning of static and dynamic brain functional networks for early mci detection. IEEE Trans. Med. Imaging pp. 39:478--87 (2020)

\bibitem{refcompicpr1}
Kam, T.E., Zhang, H., Shen, D.: A novel deep learning framework on brain functional networks for early mci diagnosis. Medical image computing and computer assisted intervention – (MICCAI 2018). pp. 293--301 (2018)

\bibitem{ph2}
Khasawneh, F.A., Munch, E.: Chatter detection in turning using persistent homology. Mechanical Systems and Signal Processing,  \textbf{70},  527–541 (2016)

\bibitem{2012-Lee-Persistent-Homology-Brain-Networks}
Lee, H., Kang, H.e.: Persistent brain network homology from the perspective of dendrogram. IEEE transactions on medical imaging  \textbf{31}(12),  2267--2277 (2012)

\bibitem{refcompicpr3}
Lee, J., KoW, Kang, E.e.a.: A unified framework for personalized regions selection and functional relation modeling for early mci identification. Neuroimage  (2021)

\bibitem{2016-Merelli-EEG}
Merelli, E., Piangerelli, M.e.: A topological approach for multivariate time series characterization: the epileptic brain. In: In Proc. EAI International Conference on Bio-inspired Information and Communications Technologies. pp. 201--204 (2016)

\bibitem{morley2011anticholinergic}
Morley, J.E.: Anticholinergic medications and cognition. Journal of the American Medical Directors Association  \textbf{12}(8),  543--543 (2011)

\bibitem{sw1}
Perea, J.A., Harer, J.: Sliding windows and persistence: An application of topological methods to signal analysis. Foundations of Computational Mathematics  \textbf{15},  799--838 (2015)

\bibitem{2021-Stolz-schizophrenia-experiments}
Stolz, B.J., Emerson, T.e.: Topological data analysis of task-based fmri data from experiments on schizophrenia. Journal of Physics: Complexity  \textbf{2}(3),  035006 (2021)

\bibitem{unknown}
Wang, F., Kapse, S., Liu, S., Prasanna, P., Chen, C.: Topotxr: A topological biomarker for predicting treatment response in breast cancer (05 2021)

\bibitem{refcompicpr4}
Wang, M., Lian, C., Yao, D.e.a.: Spatial-temporal dependency modeling and network hub detection for functional mri analysis via convolutional-recurrent network. IEEE Trans Biomed Eng pp. 2241--2252 (2020)

\bibitem{refcompicpr5}
Yang, P.e.a.: Fused sparse network learning for longitudinal analysis of mild cognitive impairment. IEEE Transactions on Cybernetics p. 233–246 (2021)

\bibitem{refcomp4}
Yang, P., Zhou, F.e.: Fused sparse network learning for longitudinal analysis of mild cognitive impairment. IEEE transactions on cybernetics  \textbf{51}(1),  233--246 (2019)

\end{thebibliography}
\end{document}